\documentclass[11pt,a4paper]{article}
\usepackage[hyperref]{acl2017}
\usepackage{times}
\usepackage{latexsym}

\usepackage{url}

\aclfinalcopy 

\def\aclpaperid{24X-J8F8H7G3C2} 

\title{PunFields at SemEval-2017 Task 7: Employing Roget's Thesaurus in Automatic Pun Recognition and Interpretation}

\author{Elena Mikhalkova \\
  Institute of Philology \\
  and Journalism \\
  Tyumen State University \\
  Tyumen, Russia, 625003 \\
  {\tt e.v.mikhalkova@utmn.ru} \\\And
  Yuri Karyakin \\
  Institute of Mathematics \\
  and Computer Science \\
  Tyumen State University \\
  Tyumen, Russia, 625003 \\
  {\tt y.e.karyakin@utmn.ru} \\}

\date{}

\begin{document}
\maketitle
\begin{abstract}
  The article describes a model of automatic interpretation of English puns, 
  based on Roget's Thesaurus. In a pun, the algorithm discovers two groups of words,
  which belong to two main semantic fields. The fields become a semantic vector, 
  based on which, an SVM classifier learns to recognize puns. A rule-based model 
  is then applied for recognition of intentionally ambiguous (target) words
  and their definitions.
\end{abstract}

\section{Introduction}

The following terminology is basic in our research of puns. {\bf A pun} is a) a short humorous genre,
where a word or phrase is intentionally used in two meanings, b) a means of expression, the essence of which 
is to use a word or phrase so that in the given context the word or phrase can be understood in two meanings simultaneously.
{\bf A target word} is a word, that appears in two meanings. {\bf A homographic pun} is a pun that 
``exploits distinct meanings of the same written word''~\cite{miller2015automatic} (these can be meanings of a polysemantic word, 
or homonyms, including homonymic word forms). {\bf A heterographic pun} is a pun, in which the target word 
resembles another word, or phrase in spelling; we will call the latter {\bf the second target word}. 
Consider the following example (the Banker joke):

\begin{quote}
``I used to be a banker, but I lost interest.''
\end{quote}

The Banker joke is a homographic pun; {\em interest} is the target word. Unlike it, the Church joke below is a heterographic pun; 
{\em propane} is the target word, {\em profane} is the second target word:

\begin{quote}
``When the church bought gas for their annual barbecue, proceeds went from the sacred to the propane.''
\end{quote} 

Our model of automatic pun analysis is based on the following premise: in a pun, there are two groups of words, 
and their meanings, that indicate the two meanings in which the target word is used. These groups can overlap, 
i.e. contain the same polysemantic words, used in different meanings. 

In the Banker joke, words, and collocations {\em banker}, {\em lost interest} point at the professional status of the narrator, and his/her 
career failure. At the same time, {\em used to}, {\em lost interest} tell a story of losing emotional attachment to the profession: 
the narrator became disinterested. The algorithm of pun recognition, which we suggest, discovers these two groups of words, based on common 
semes\footnote{Bits of meaning. Semes are some parts of meaning, present both in the word and in its hypernym.
Moving up the taxonomy, like Thesaurus, or WordNet, hypernyms become more general, and the seme, connecting them to the word, becomes more general, too.} 
(Subtask 1), finds the words, which belong to the both groups, and chooses the target word (Subtask 2), 
and, based on the common semes, picks up the best suitable meaning, which the target word exploits (Subtask 3). 
In case of heterographic puns, in Subtask 2, the algorithm looks for the word, or phrase, which appears in one group 
and {\em not} in the other.

\section{Subtask 1: Mining Semantic Fields}

We will call a semantic field a group of words and collocations, which share a common seme. In taxonomies, like WordNet~\cite{kilgarriff2000wordnet}, and 
Roget's Thesaurus~\cite{roget2004roget} (further referred to as Thesaurus), semes appear as hierarchies of word meanings. Top-levels
attract words with more general meanings (hypernyms). For example, Thesaurus has six top-level Classes, that divide into Divisions, that divide
into Sections, and so on, down to the fifth lowest level. WordNet's structure is not so transparent. CITE!!! 10 TOP-semes 
Applying such dictionaries to get semantic fields (the mentioned common groups of words) in a pun is, therefore, the task of finding 
two most general hypernyms in WordNet, or two relevant Classes among the six Classes in Thesaurus. We chose Thesaurus, as its structure 
is only five levels deep, Classes labels are not lemmas themselves, but arbitrary names (we used numbers instead), and it allows parsing 
on a certain level, and insert corrections (adding lemmas, merging subsections, etc.\footnote{For example, we edited Thesaurus, adding words, 
which were absent in it. If a word in a pun was missing in Thesaurus, the system checked up for its hypernyms in Wordnet, and added the word 
to those Sections in Thesaurus, which contained the hypernyms.}). After some experimentation, instead of Classes, we chose to search for relevant 
Sections, which are 34 subdivisions of the six Classes\footnote{Sections are not always {\em immediate} subdivisions of a Class. 
Some Sections are grouped in Divisions.}.

After normalization (including change to lowercase; part-of-speech tagging, tokenization, and lemmatization with NLTK tools~\cite{bird2009natural}; 
collocation extraction\footnote{To extract collocations and search for them in Thesaurus, we applied our own procedure, 
based on a part-of-speech analysis.}; stop-words removal\footnote{After lemmatization, all words are analyzed in collocations, but only nouns, adjectives, and
verbs compose a list of separate words.}), the algorithm collects Section numbers for every word, and collocation, and removes duplicates 
(in Thesaurus, homonyms proper can belong to different subdivisions in the same or different Sections). 
Table~\ref{tab:sem} shows what Sections words of the Banker joke belong to.

\begin{table*}
\centering
\begin{tabular}{lll}
  \bf Word & \bf Section No., Section name in Thesaurus \\ 
  \hline
  I & - \\
  use & 24, Volition In General \\
  & 30, Possessive Relations \\
  to & - \\
  be & 0, Existence \\
  &  19, Results Of Reasoning \\
  a & - & \\
  banker & 31, Affections In General \\
  &  30, Possessive Relations \\
  but & - \\
  lose & 21, Nature Of Ideas Communicated \\
  & 26, Results Of Voluntary Action \\
  & 30, Possessive Relations \\
  & 19, Results Of Reasoning \\
  interest & 30, Possessive Relations \\
  & 25, Antagonism \\
  & 24, Volition In General \\
  & 7, Causation \\
  & 31, Affections In General \\
  & 16, Precursory Conditions And Operations \\
  & 1, Relation \\
\end{tabular}
\caption{Semantic fields in the Banker joke}\label{tab:sem}
\end{table*}

Then the semantic vector of a pun is calculated. Every pun \(p\) is a vector in a 34-dimensional space:
\[p_{i}=p_{i}(s_{1i}, s_{2i},...,s_{34i})\]\label{f:pi}
The value of every element \(s_{ki}\) equals the number of words in a pun, which belong to a Section \(S_{k}\).
The algorithm passes from a Section to a Section, each time checking every word \(w_{ji}\) in the bunch of extracted words \(l_{i}\). 
If a word belongs to a Section, the value of \(s_{ki}\) RAISES BY???? 1:
\[s_{ki}=\displaystyle\sum_{j=1}^{l_{i}} \left\{1\middle|w_{ji} \in S_{k}\right\}, k=1,2,...,34, i=1,2,3...\]\label{f:ski}
For example, the semantic vector of the Banker joke looks as follows: see Table~\ref{tab:pbnker}.
\begin{table*}
\centering
\begin{tabular}{lll}
  \(p_{Banker}\) &  \(\{ 1 , 1 , 0 , 0 , 0 , 0 , 0 , 1 , 0 , 0 , 0 , 0 , 0 , 0 , 0 , 0 , 1 , 0 , 0 , 2 , 0 , 1 , 0 , 0 , 2 , 1 , 1 , 0 , 0 , 0 , 4 , 2, 0, 0 \}\)\\ 
\end{tabular}
\caption{Semantic vector of the Banker joke}\label{tab:pbnker}
\end{table*}

To test the algorithm, we, first, collected 2484 puns from different Internet resources and, second, built a corpus
of 2484 random sentences of length 5 to 25 words from different NLTK corpora~\cite{bird2009natural} plus several hundred aphorisms and 
proverbs from different Internet sites. We shuffled and split the sentences into two equal groups, the first two forming 
a training set, and the other two a test set. The classification was conducted, using different Scikit-learn~\cite{pedregosa2011scikit} 
algorithms. We also singled out 191 homographic puns, and 198 heterographic puns, and tested them against the same number of random
sentences.
In all the tests\footnote{The tests were run before the competition. Results of the competition for our system are given in Table~\ref{results}.}, 
the Scikit-learn algorithm of SVM with the Radial Basis Function (RBF) kernel produced the highest average 
F-measure results (\(\bar{f}=\frac{f_{puns}+f_{random}}{2}\)). In addition, its results are smoother, comparing the difference 
between precision, and recall (which leads to the highest F-measure scores) within the two classes (puns, and random sentences), 
and between the classes (average scores). Table~\ref{punrec} illustrates results of different algorithms in class ``Puns'' (not 
average results between puns, and not puns). The results were higher 
for the split selection, reaching 0.79 (homographic), and 0.78 (heterographic) scores of F-measure. The common selection 
got the maximum of 0.7 for average F-measure in several tests. The higher results of split selection may be due to 
a larger training set.

\begin{table*}
\centering
\begin{tabular}{llll}
\hline
\bf Method & \bf Precision & \bf Recall & \bf F-measure \\
\hline
\bf Common selection \\
\hline
SVM with linear kernel & 0.67 & 0.68 & 0.67 \\
SVM with polynomial kernel & 0.65 & 0.79 & 0.72 \\
SVM with Radial Basis Function (RBF) kernel & 0.70 & 0.70 & 0.70 \\
SVM with linear kernel, normalized data & 0.62 & 0.74 & 0.67 \\
\hline
\bf Homographic puns \\
\hline
SVM with RBF kernel & 0.79 & 0.80 & 0.79 \\
Multinomial Naive Bayes & 0.71 & 0.80 & 0.76 \\
Logistic Regression, standardized data & 0.77 & 0.71 & 0.74 \\
\hline
\bf Heterographic puns \\
\hline
SVM with RBF kernel & 0.77 & 0.79 & 0.78 \\
Logistic Regression & 0.74 & 0.75 & 0.74 \\
\hline
\end{tabular}
\caption{Tests for pun recognition.}\label{punrec}
\end{table*}

\section{Subtask 2: Hitting the Target Word}

We suggest that, in a homographic pun, the target word is a word, which immediately belongs to two semantic fields; 
in a heterographic pun, the target word belongs to at least one discovered semantic field, and does not belong to the other. 
However, in reality, words in a sentence tend to belong to too many fields, and they create noise in the search. 
To reduce influence of noisy fields, we included such non-semantic features in the model as
the tendency of the target word to occur at the end of a sentence, and part-of-speech distribution, given in~\cite{miller2015automatic}.
A-group (\(W_{A}\)) and B-group (\(W_{B}\)) are groups of words in a pun, which belong to the two semantic fields, 
sharing the target word. Thus, for some \(s_{ki}\), \(k\) becomes \(A\), or \(B\)~\footnote{\(s_{ki}\) is always an integer; \(W_{A}\) and \(W_{B}\) 
are always lists of words; \(A\) is always an integer, \(B\) is a list of one or more integers.}. A-group attracts the maximum number of words in a pun:
\[s_{Ai}=\max_{k} s_{ki}, k=1,2,...,34 \]

In the Banker joke, \(s_{Ai}=4, A=30\) (Possessive Relations); words, that belong to this group, are {\em use}, {\em lose}, 
{\em banker}, {\em interest}. B-group is the second largest group in a pun:
\[s_{Bi}=\max_{k} ( s_{ki}/s_{Ai} ), k=1,2,...,34 \]
In the Banker joke, \(s_{Bi}=2\). There are three groups of words, which have two words in them: \(B_{1}=19\), 
Results Of Reasoning: {\em be}, {\em lose}; \(B_{2}=24\), Volition In General: {\em use}, {\em interest}; 
\(B_{3}=31\), Affections In General: {\em banker}, {\em interest}. Ideally, there should be a group of about three words, 
and collocations, describing a person`s inner state ({\em used to be}, {\em lose}, {\em interest}), and two words 
({\em lose}, {\em interest}) in \(W_{A}\) are a target phrase. However, due to the shortage of data 
about collocations in dictionaries, \(W_{B}\) is split into several smaller groups. Consequently, to find the target word, 
we have to appeal to other word features. In testing the system on homographic puns, we relied on the polysemantic character of words.
If in a joke, there are more than one value of \(B\), \(W_{B}\) candidates merge into one, with duplicates removed, 
and every word in \(W_{B}\) becomes the target word candidate: \(c \in W_{B}\). In the Banker joke, \(W_{B}\) is a list of 
{\em be}, {\em lose}, {\em use}, {\em interest}, {\em banker}; \(B=\{19,24,31\}\). Based on the definition of the target word 
in a homographic pun, words from \(W_{B}\), that are also found in \(W_{A}\), should have a privilege. Therefore, 
the first value \(v_{\alpha}\), each word gets, is the output of the Boolean function:
\[v_{\alpha}(c)=\left\{
 \begin{array}{lr}
  2 \quad \mathrm{if } (c \in W_{A})\wedge(c \in W_{B}) \\
  1 \quad \mathrm{if } (c \notin W_{A})\wedge(c \in W_{B}) \\
 \end{array}
\right.
\]

The second value \(v_{\beta}\) is the absolute frequency of a word in the union of \(B_{1}\), \(B_{2}\), etc., 
including duplicates: \(v_{\beta}(c)=f_c(W_{B_{1}} \cup W_{B_{2}} \cup W_{B_{3}})\).

The third value \(v_{\gamma}\) is a word position in the sentence: the closer the word is to the end, the
bigger this value is. If the word occurs several times, the algorithm counts the average of the
sums of position numbers.

The fourth value is part-of-speech probability \(v_{\delta}\). Depending on the part of speech, the word
belongs to, it gets the following rate:
\[v_{\delta}(c) = \left\{
  \begin{array}{lr}
    0.502, if c - Noun \\
    0.338, if c - Verb \\
    0.131, if c - Adjective \\
    0.016, if c - Adverb \\
    0.013, otherwise \\
  \end{array}
\right.
\]

The final step is to count rates, using multiplicative convolution, and choose the word with the
maximum rate:
\[z_{1}(W_{B})=\left\{c|\max_{c} (v_{\alpha} \times v_{\beta} \times v_{\gamma} \times v_{\delta}) \right\}\]

Values of the Banker joke are illustrated in Table~\ref{valB}.

\begin{table}
\centering
\small
\begin{tabular}{cccccc}
\begin{tabular}{|l|l|l|l|l|l|}
\hline
\bf Word form & \(v_{\alpha}\) & \(v_{\beta}\) & \(v_{\gamma}\) & \(v_{\delta}\) & \(v_{W_{Bk}}\) \\
\hline
be & 1 & 1 & 4 & 0.338 & 1.352 \\
lose & 2 & 1 & 9 & 0.338 & 6.084 \\
use & 2 & 1 & 2 & 0.338 & 1.352 \\
interest & 2 & 2 & 10 & 0.502 & \bf 20.08 \\
banker & 2 & 1 & 6 & 0.502 & 6.024 \\
\hline
\end{tabular}
\end{tabular}
\caption{Values of the Banker joke.}\label{valB}
\end{table}

In the solution for heterographic puns, we built a different model of B-group. Unlike homographic
puns, here the target word is missing in \(W_{B}\) (the reader has to guess the word or phrase,
homonymous to the target word). Accordingly, we rely on the completeness of the union of \(W_{A}\)
and \(W_{B}\): among the candidates for \(W_{B}\) (the second largest groups), such groups are relevant,
that form the longest list with \(W_{A}\) (duplicates removed). In Ex. 2 (the Church joke), \(W_{A}={go, gas,
annual, barbecue, propane}\), and two groups form the largest union with it: \(W_{B}={buy,
proceeds} + {sacred, church}\). Every word in \(W_{A}\) and \(W_{B}\) can be the target word. The privilege
passes to words, used only in one of the groups. Ergo, the first value is:
\[v_{\alpha}(c)=\left\{
 \begin{array}{lr}
  2 \quad \mathrm{if } (c \in W_{A})\oplus(c \in W_{B}) \\
  1 \quad \mathrm{otherwise} \\
 \end{array}
\right.
\]
Frequencies are of no value here; values of position in the sentence, and part-of-speech
distribution remain the same. The function output is:
\[z_{1}(W_{B})=\left\{c|\max_{c} (v_{\alpha} \times v_{\gamma} \times v_{\delta}) \right\}\]

Values of the Church joke are illustrated in Table~\ref{valC}.
\begin{table}
\centering
\small
\begin{tabular}{ccccc}
\begin{tabular}{|l|l|l|l|l|}
\hline
\bf Word form & \(v_{\alpha}\) & \(v_{\gamma}\) & \(v_{\delta}\) & \(v_{W_{Ak}}, v_{W_{Bk}}\) \\
\hline
propane & 2 & 18 & 0.502 & \bf 18.072 \\
annual & 2 & 8 & 0.131 & 2.096 \\
gas & 2 & 5 & 0.502 & 5.02 \\
sacred & 2 & 15 & 0.338 & 10.14 \\
church & 2 & 3 & 0.502 & 3.012 \\
barbecue & 2 & 9 & 0.502 & 9.036 \\
go & 2 & 12 & 0.338 & 8.112 \\
proceeds & 2 & 11 & 0.502 & 11.044 \\
buy & 2 & 4 & 0.338 & 2.704 \\
\hline
\end{tabular}
\end{tabular}
\caption{Values of the Church joke.}\label{valC}
\end{table}

\section{Subtask 3: Mapping Roget's Thesaurus to Wordnet}
In the last phase, we implemented an algorithm which maps Roget's Sections to synsets in Wordnet. 
In homographic puns, definitions of a word in Wordnet are analyzed similarly to words in a pun, 
when searching for semantic fields, the words belong to. For example, words from the definitions 
of the synset {\em interest} belong to the following Roget's Sections: 
Synset(interest.n.01)=a sense of concern with and curiosity about someone or something: 
(21, 19, 31, 24, 1, 30, 6, 16, 3, 31, 19, 12, 2, 0); Synset(sake.n.01)=a reason for wanting something done: 
15, 24, 18, 7, 19, 11, 2, 31, 24, 30, 12, 2, 0, 26, 24, etc. When A-Section is discovered (for example,
in the Banker joke, A=30 (Possessive Relations)), the synset with the maximum number of words in its definition,
which belong to A-Section, becomes the A-synset. The B-synset is found likewise for the B-group with the exception 
that it should not coincide with A-synset. In heterographic puns the B-group is also a marker of the second target word. 
Every word in the index of Roget's Thesaurus is compared to the known target word using Damerau-Levenshtein distance. 
The list is sorted in increasing order, and the algorithm begins to check what Roget's Sections every word belongs to, 
until it finds the word that belongs to a Section (or the Section, if there is only one) in the B-group. This word becomes 
the second target word.

Nevertheless, as we did not have many trial data, but for the four examples, released before the competition, the first trials 
of the program on a large collection returned many errors, so we changed the algorithm for the B-group as follows.

{\em Homographic puns, first run.} B-synset is calculated on the basis of sense frequencies 
(the output is the most frequent sense). If it coincides with A-synset, the program returns the second frequent synset.

{\em Homographic puns, second run.} B-synset is calculated on the basis of Lesk distance, using
built-in NLTK Lesk function~\cite{bird2009natural}. If it coincides with A-synset, the program returns another synset on the basis of sense frequencies, 
as in the first run.

{\em Heterographic puns, first run.} The second target word is calculated, based on Thesaurus and Damerau-Levenshtein distance; 
words, missing in Thesaurus, are analyzed as their WordNet hypernyms. In both runs for heterographic puns, synsets are
calculated, using the Lesk distance.

{\em Heterographic puns, second run.} The second target word is calculated on the basis of Brown corpus (NLTK~\cite{bird2009natural}): 
if the word stands in the same context in Brown as it is in the pun, it becomes the target word. The size of the 
context window is (0; +3) for verbs, (0;+2) for adjectives; (-2;+2) for nouns, adverbs and other parts of speech 
within the sentence, where a word is used.

Table~\ref{results} illustrates competition results of our system.

\begin{table}
\centering
\small
\begin{tabular}{ccccc}
\begin{tabular}{|l|l|l|l|l|}
\hline
\bf Task & \bf Precision & \bf Recall & \bf Accuracy & \bf F1 \\
\hline
1, Ho.\footnote{Homographic.} & 0.8019 & 0.7785 & 0.7044 & 0.7900 \\
1, He.\footnote{Heterographic.} & 0.7585 & 0.6326 & 0.5938 & 0.6898 \\
\hline
\bf Task & \bf Coverage & \bf Precision & \bf Recall & \bf F1 \\
\hline
2, Ho., run 1 & 1.0000 & 0.3279 & 0.3279 & 0.3279 \\
2, Ho., run 2 & 1.0000 & 0.3167 & 0.3167 & 0.3167 \\
2, He., run 1 & 1.0000 & 0.3029 & 0.3029 & 0.3029 \\ 
2, He., run 2 & 1.0000 & 0.3501 & 0.3501 & 0.3501 \\
3, Ho., run 1 & 0.8760 & 0.0484 & 0.0424 & 0.0452 \\
3, Ho., run 2 & 1.0000 & 0.0331 & 0.0331 & 0.0331 \\
3, He., run 1 & 0.9709 & 0.0169 & 0.0164 & 0.0166 \\ 
3, He., run 2 & 1.0000 & 0.0118 & 0.0118 & 0.0118 \\
\hline
\end{tabular}
\end{tabular}
\caption{Competition results.}\label{results}
\end{table}

\section{Conclusion}

The system, that we introduced, is based on one general supposition about the semantic structure of puns and combines two
types of algorithms: supervised learning and rule-based. Not surprisingly, the supervised learning algorithm showed 
better results in solving an NLP-task, than the rule-based.
Also, in this implementation, we tried to combine two very different dictionaries (Roget's Thesaurus and Wordnet). And,
although reliability of Thesaurus in reproducing a universal semantic map can be doubted, it seems to be a quite effective
source of data, still, when used in Subtask 1. The attempts to map it to Wordnet seem rather weak, so far, concerning
the test results, which also raises a question: if different dictionaries treat meaning of words differently, can there be an
objective and/or universal semantic map, to apply as the foundation for any WSD task? 

\bibliography{Mikhalkova_ev}

\begin{thebibliography}{}
\expandafter\ifx\csname natexlab\endcsname\relax\def\natexlab#1{#1}\fi

\bibitem[{Bird et~al.(2009)Bird, Klein, and Loper}]{bird2009natural}
Steven Bird, Ewan Klein, and Edward Loper. 2009.
\newblock {\em Natural language processing with {P}ython: analyzing text with
  the natural language toolkit\/}.
\newblock O'Reilly Media, Inc.

\bibitem[{Kilgarriff and Fellbaum(2000)}]{kilgarriff2000wordnet}
Adam Kilgarriff and Christiane Fellbaum. 2000.
\newblock Wordnet: {A}n electronic lexical database.

\bibitem[{Miller and Gurevych(2015)}]{miller2015automatic}
Tristan Miller and Iryna Gurevych. 2015.
\newblock Automatic disambiguation of {E}nglish puns.
\newblock In {\em ACL (1)\/}. pages 719--729.

\bibitem[{Pedregosa et~al.(2011)Pedregosa, Varoquaux, Gramfort, Michel,
  Thirion, Grisel, Blondel, Prettenhofer, Weiss, Dubourg
  et~al.}]{pedregosa2011scikit}
Fabian Pedregosa, Ga{\"e}l Varoquaux, Alexandre Gramfort, Vincent Michel,
  Bertrand Thirion, Olivier Grisel, Mathieu Blondel, Peter Prettenhofer, Ron
  Weiss, Vincent Dubourg, et~al. 2011.
\newblock Scikit-learn: Machine learning in {P}ython.
\newblock {\em Journal of Machine Learning Research\/} 12(Oct):2825--2830.

\bibitem[{Roget(2004)}]{roget2004roget}
Peter~Mark Roget. 2004.
\newblock Roget's thesaurus of {E}nglish words and phrases.
\newblock Project Gutenberg.

\end{thebibliography}
\bibliographystyle{acl_natbib}

\end{document}